# EXPLORING LANGUAGE SIMILARITIES WITH DIMENSIONALITY REDUCTION TECHNIQUES

Sangarshanan Veeraraghavan
Final Year Undergraduate
VIT Vellore
v.sangarshanan2015@vit.ac.in

## Abstract

In recent years several novel models were developed to process natural language, development of accurate language translation systems have helped us overcome geographical barriers and communicate ideas effectively. These models are developed mostly for a few languages that are widely used while other languages are ignored. Most of the languages that are spoken share lexical, syntactic and sematic similarity with several other languages and knowing this can help us leverage the existing model to build more specific and accurate models that can be used for other languages, so here I have explored the idea of representing several known popular languages in a lower dimension such that their similarities can be visualized using simple 2 dimensional plots. This can even help us understand newly discovered languages that may not share its vocabulary with any of the existing languages.

## 1. Introduction

Language is a method of communication and ironically has long remained a communication barrier. Written representations of all languages look quite different but inherently share similarity between them. For example if we show a person with no knowledge of English alphabets a text in English and then in Spanish they might be oblivious to the similarity between them as they look like gibberish to them. There might also be several languages which may seem completely different to even experienced linguists but might share a subtle hidden similarity as they is a possibility that languages with no shared vocabulary might still have some similarity.

Lexical similarity between languages are fairly easy to determine but it depends purely on the vocabulary of the language and not how they are structured to form a sentence. Hence this method might not be applicable to languages with no common words.

Languages have evolved over several years and if we were to establish concrete relationships among them it would become fairly easy to develop language specific models for lesser known languages which would be much more accurate than generic models.

## 2. Proposed approach

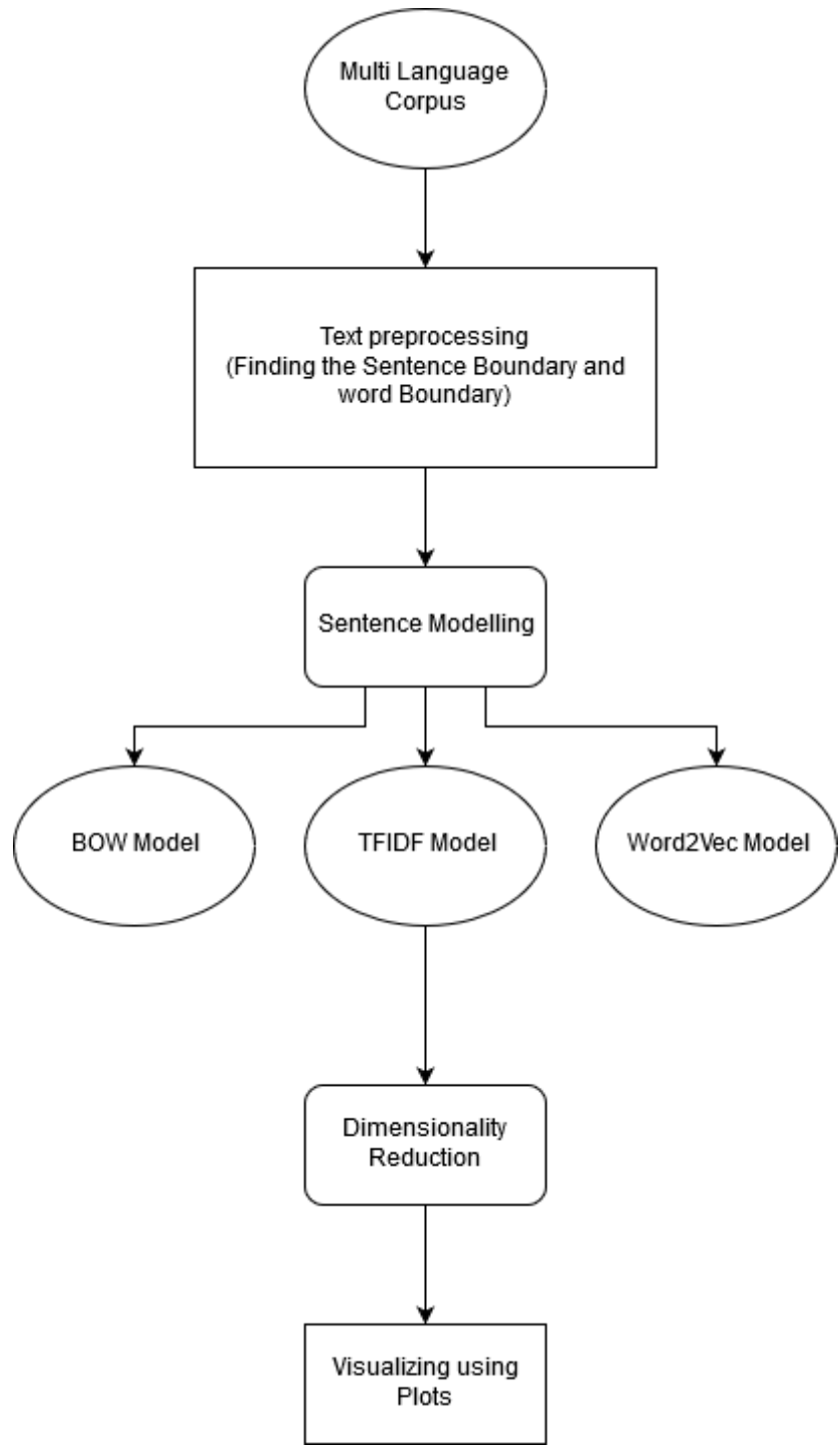

## 2.1 Text Preprocessing

Since our corpus is made of different languages, it is quite difficult to use a single type of preprocessing method. The difference in preprocessing methods arise due to the different writing systems. Languages can be classified based into three types based on their writing system

1. Logographic (Every symbol is a morpheme)
2. Syllabic (Every symbol is a syllable)
3. Alphabetic (Every symbol is a Phoneme)

We can separately identify the sentences based on the writing system and remove the unnecessary punctuations. Stopwords should be kept intact as sometimes they give us critical information on the language (Since stopwords are language specific)

For Tokenization we can use the standard stanford word segmenter or the weiba segmentation algorithm, for Chinese whereas we can use Mecab for Japanese and UETsegmenter for Vietnamese. We can use nltk for Arabic and for Latin based languages and many Indian languages, words are space separated and hence tokenization is much easier

The final goal is split the documents into sentences and then into characters after which we can group them corresponding to their languages.

## 2.2 Sentence Modelling

Sentence Modelling is a way to represent sentences as vectors. We compute the vector representation of each word and find their weighted average using PCA

Calculating the vector representation of the words in a sentence is called as Word embedding. There are several ways to do this and it all depends on the task and the dataset. We can use either a pre trained embedding or train our own one. For this task as I am going to working with several lesser known languages and these might not have an already trained vector representation so it is better to train our own embedding.

A bag of words model can be achieved using the CountVectorizer() function provided by Sklearn which can be used to calculate word vectors by fitting it on our data and transforming it into vectors. The encoded vector is returned with a length of the entire vocabulary and an integer count for the number of times each word appeared in the document.

TF-IDF or Term frequency inverse document frequency vectorizer can be used to calculate the vectors corresponding to the words as it can tokenize the documents and learn the vocabulary and the inverse document frequency weights which would help in encoding new documents.

We can use TfidfVectorizer() function provided by Sklearn to calculate the vectors. We initially use the fit() function to learn the vocabulary from our corpus followed by transform to use the knowledge to encode our sentences corresponding to the tfidf values but the vector that we get is sparse and will contain plenty to zeros given the extent of our corpus, this can be handled by calling the toarray() function which converts the sparse vectors into a numpy array.

Sentence embedding can be a combination of word embedding of all words in a sentence. Using bag-of-words or one-hot encoding models for embedding might seem simple and viable as our task does not involve complicated deep neural networks but in order to capture the Syntactic (Structure) and Semantic (meaning) relationship between the sentences corresponding to different languages we cannot rely on these naive methods.

Word2vec is a two layer neural network that can also be used to generate robust word vectors and these vectors can reconstruct the linguistic context of the word by leveraging a skip gram model

## 2.3 Dimensionality Reduction

Projecting a higher dimensional data into two dimensions makes it easier to understand and visualize. This can be applied to our data so that we would be able to observe the relationship between the languages by reducing the dimensions of the vector that we obtained from the previous step.

For this task I chose Uniform Manifold Approximation and projection for dimensionality reduction. UMAP uses local manifold approximations and patches together their local fuzzy simplicial set representations which would construct a topological representation of the given higher dimensional data. It is superior to t-sne in terms of visualization quality and also preserves more of the global structure with superior runtime.

UMAP has a topological foundation makes it feasible for larger data sets and when we are handling scripts from several languages we need significantly higher visualization quality that can be provided by UMAP.

Dimensionality reduction can lead to loss of information that can be represented better in higher dimension but with higher dimension it is very difficult to make sense of the data. The two categories in in dimensionality reduction are feature selection and feature extraction. Feature selection is used to identify a subset of features that would lead to minimal loss in information whereas feature extraction refers to using techniques like

Principal component analysis, linear discriminant analysis and several other methods to transform the higher dimensional data into lower dimension.

## 3. Exploring the Languages

Plotting all the existing language corpuses on a 2D graph will get clumsy and pinpointing language similarities would become a tedious process and hence we must be careful in picking the languages that we are plotting. We can analyse languages based on their families, roots, dialects, origin or several other factors that may influence similarity

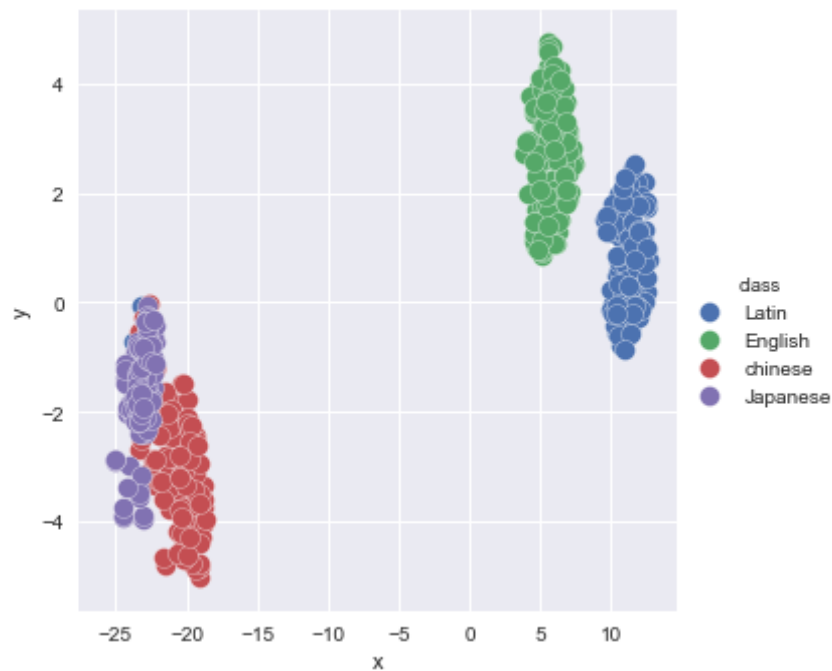

Fig-1 Plotting Latin and English along with Chinese and Japanese corpuses (Tfidf vectors)

In the above graph it is very easy to notice two clusters ['Latin','English'] and ['Chinese','japanese']

English vocabulary draws heavily from Latin even though it is a Germanic language and hence Latin and English are clustered together whereas Japanese and Chinese are clustered together due to their similar writing system called 汉字. It is called Kanji in

Japanese and Hanzi in Chinese. Chinese characters were imported by japanese a long time ago and these both language share a lot of similarity in the use of characters.

Suppose I replace Japanese with Arabic and consider more European languages like Dutch and Danish to create a more interesting visualization. The Plot now includes English, Chinese, Arabic, Dutch and Danish

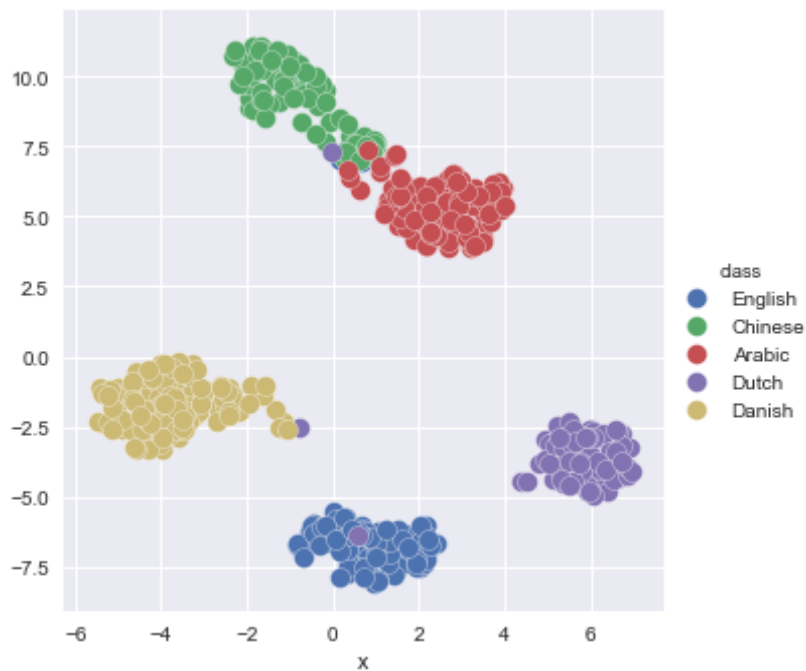

Fig 1.1 TFIDF embedding plot of Chinese-Arabic-English-Dutch-Danish

Dutch Danish and English are present in the lower half of the graph whereas Chinese and Arabic occupy the upper region of the graph. English Dutch and Danish come under the class of Germanic languages. Dutch and Danish are not mutually intelligible as Dutch is West Germanic and Danish is North Germanic but there is some commonality between these languages which also coincides with English which happens to be West Germanic along with Dutch and hence Dutch and English are closer in the plot along with some intersection between the clusters

### 3.1 Ancient world Languages

There are certain languages that are considered to be the mother of several other languages and has attained antiquity. They can be called as classical languages which are a group of ancient languages that have given birth to several other languages of the same kind. A classical language should have an independent writing system that evolved on its own without the help of any other language and also an extremely rich ancient literature to prove it.

Comparing such ancient languages can give us an idea as to how the other languages have evolved. I have considered Chinese, Greek, Sanskrit, Hebrew, Tamil, Latin and Arabic and compared them.

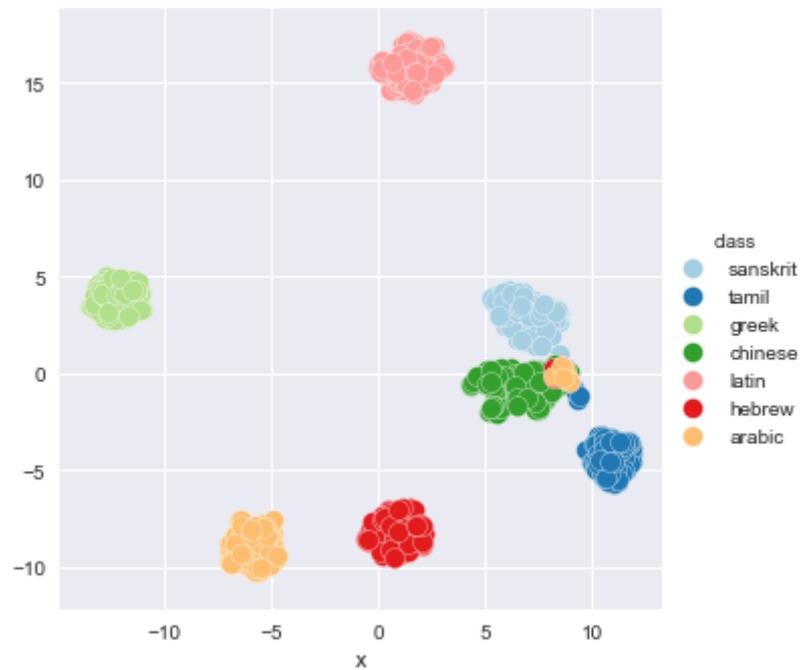

Fig-2:   Bag of words embedding

A simple bag of words embedding shows us similarity between Chinese, Sanskrit and Tamil and this might be because all three have Asian roots. Arabic and Hebrew seem to share similarities with these clusters and maybe this is because of the fact that Arabic and Hebrew share lexical similarity (about 58.5%), grammatical correspondence, and mutual intelligibility. Greek and Latin form separate clusters in the 2D space

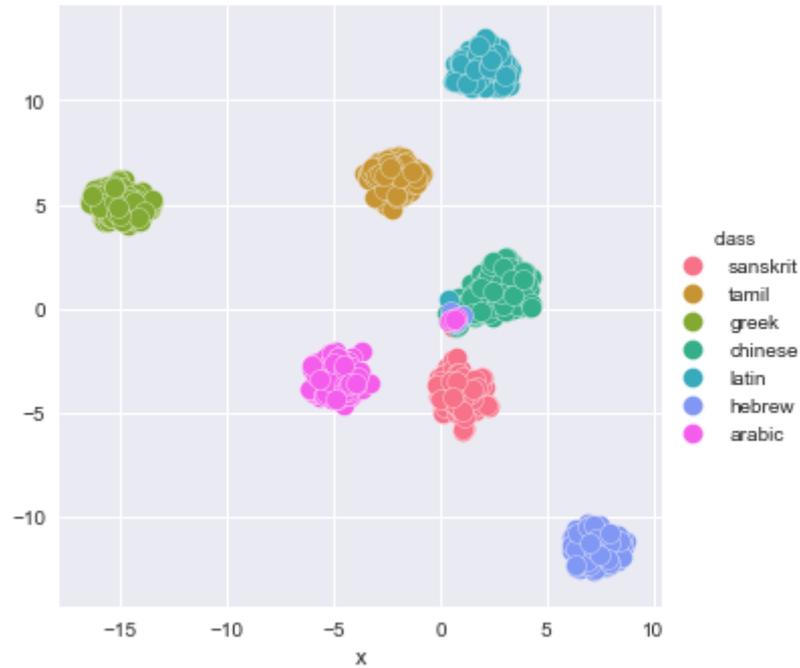

Fig3- TFIDF embedding

Even in this cluster we can see the clear overlap between Hebrew and Arabic. Both these languages come under the class of Semitic languages. These are a branch of Afro-asiatic languages that were used in the Middle East. Arabic and Hebrew flourished with the help of Islamic and Jewish scholars. Again we can see that the rest of the languages are clustered separately. An odd detail here is how both Arabic and Hebrew overlap with Chinese and the presence of Sanskrit near this group of clusters.

Both TFIDF and bag of words embedding can give us an idea of language similarity that can be quite easily detected by human even though it reduces the effort. If we need more of a neural word embedding and so we have to move to a word2vec model that leverages a two layer neural network to cluster the vector groups with high similarity together in the Vectorspace. Word2vec vectors are the distributed numeric representations of word vectors and features such as context of individual words. Word2vec can achieve this using just the corpus without human intervention

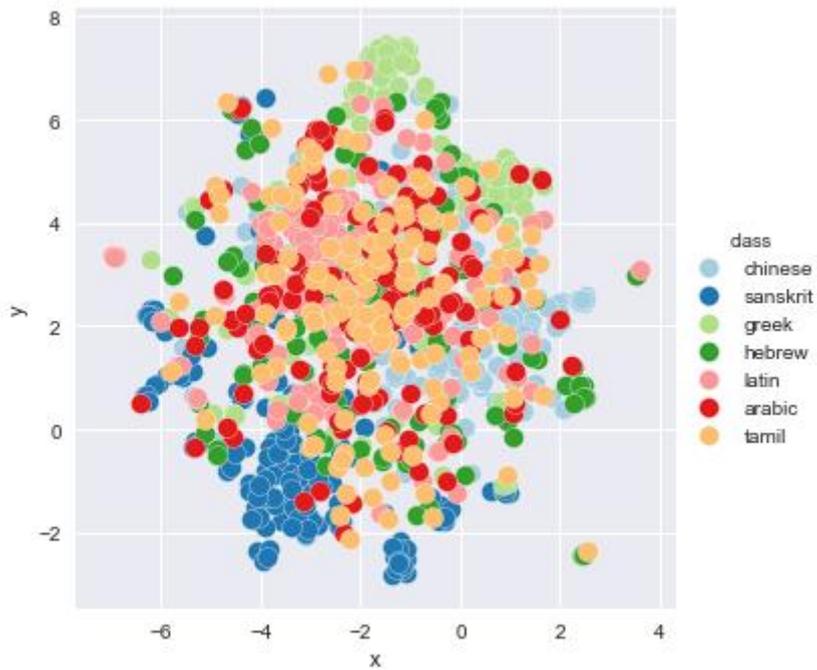

Fig 4 – Classical languages with Word2vec embedding

Bible is used as the corpus for every language and hence the vector representations of sentences tend to be similar and so unlike TFIDF and count embedding word2vec clusters all the languages together in the vector space

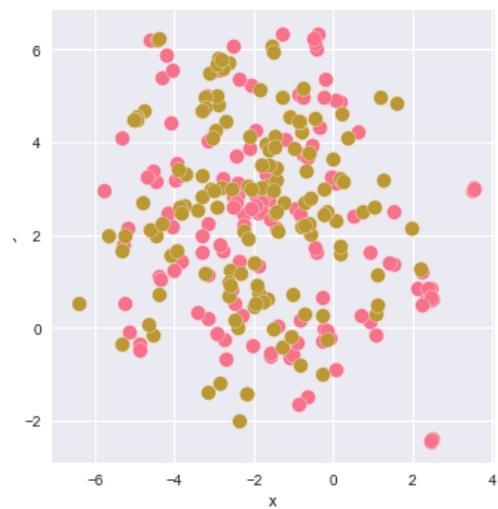
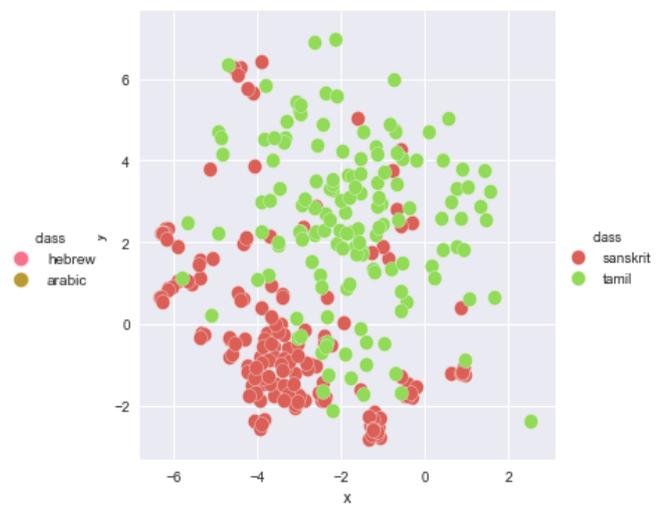

HEBREW AND ARABIC               SANSKRIT AND TAMIL

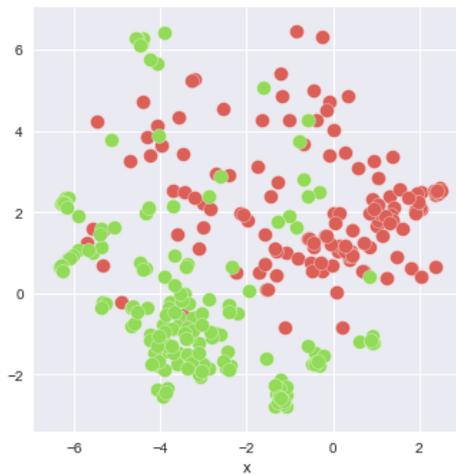 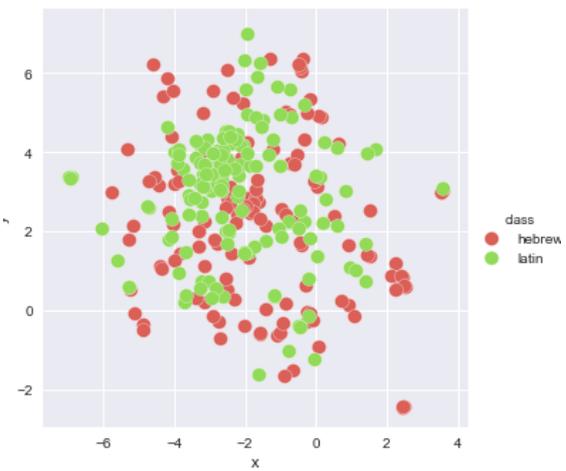

| CHINESE AND SANSKRIT | HEBREW AND LATIN |

The representation of word2vec depends on the presence of sufficient data, usage and the context to determine the meaning of a word based on its past occurrences.

In the plots we can see that languages with no common roots such as Chinese and Sanskrit followed by Sanskrit and Tamil are clustered into two even though the boundary between those clusters are hard to define. This indicates some similarity due to the use of similar corpus but in the case of languages like Hebrew and Arabic the boundary between languages doesn't seem to exist and it appears as one huge cluster. This might be because of their common origin in the Middle East.

### 3.2 Indo Aryan and Dravidian Languages

The languages spoken in the Indian Peninsula comprise of Indo Aryan languages that are used in the northern parts of the subcontinent and Dravidian languages are used in the southern part of the subcontinent. The Dravidian languages have heavily influences indo Aryan languages and this is evident in the script of Rigveda which includes several borrowed words from the Dravidian vocabulary. Over the years there has been several interactions between Indo Aryan and Dravidian languages.

Indo Aryan languages refer to the languages that were spoken by the Aryan people who moved to the Indian subcontinent in prehistoric times. The oldest Indo Aryan language is Sanskrit and it was found in the Vedic scriptures which date to 1500 BCE.

Indo Aryan languages are spoken in North-India, Pakistan, Bangladesh, Nepal, Sri Lanka, Myanmar and Maldives whereas Dravidian languages are spoken majorly in the South-India and Sri Lanka but also in countries like Malaysia and Singapore.

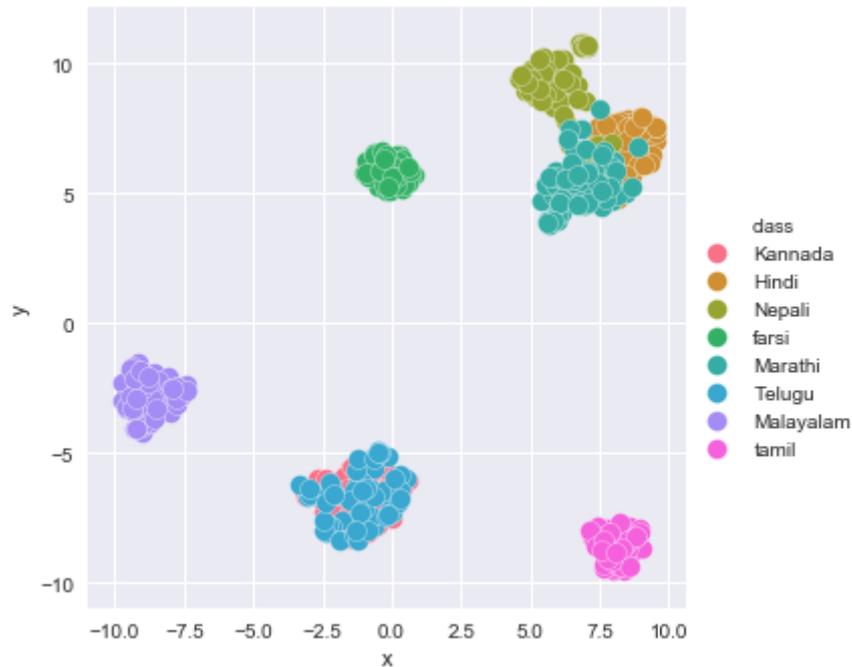

Bag of words embedding

Dravidian languages like Tamil, Kannada, Telugu and Malayalam are clustered in the bottom half of the graph whereas the Indo Aryan languages are clustered on top.

On the bottom half of the graph we can also see a complete overlap between Telugu and Kannada as both of them have their writing systems derived from the Kadamba script. The evolution of both Kannada and Telugu were heavily influenced by the Chalukya dynasty.

On the top half of the graph Farsi is clustered separately whereas Hindi, Nepali and Marathi clusters overlap with each other. The three languages clustered together follow the Devanagari script which is one of the most used and adopted writing systems in the world. Farsi has evolved from Arabic and Persian and hence is clustered separately from the languages that evolved from the Devanagari writing system.

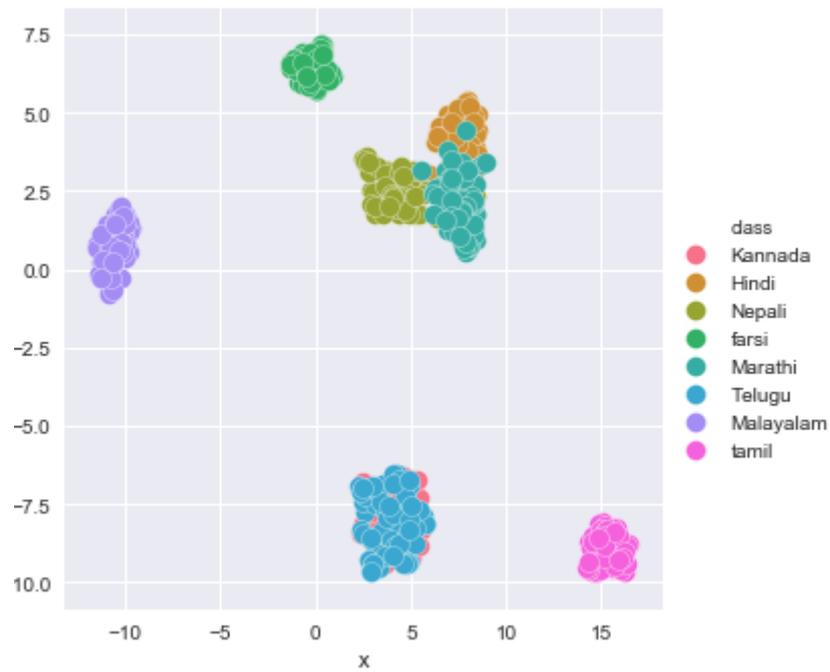

TFIDF Embedding

In TFIDF embedding Malayalam is clustered near the top. This is because Malayalam shares a high similarity with Sanskrit. Malayalam has borrowed several alphabets and grammatical rules from Sanskrit. Hence we can say that Malayalam scripts have heavily borrowed from both Dravidian scripts and Indo Aryan scripts.

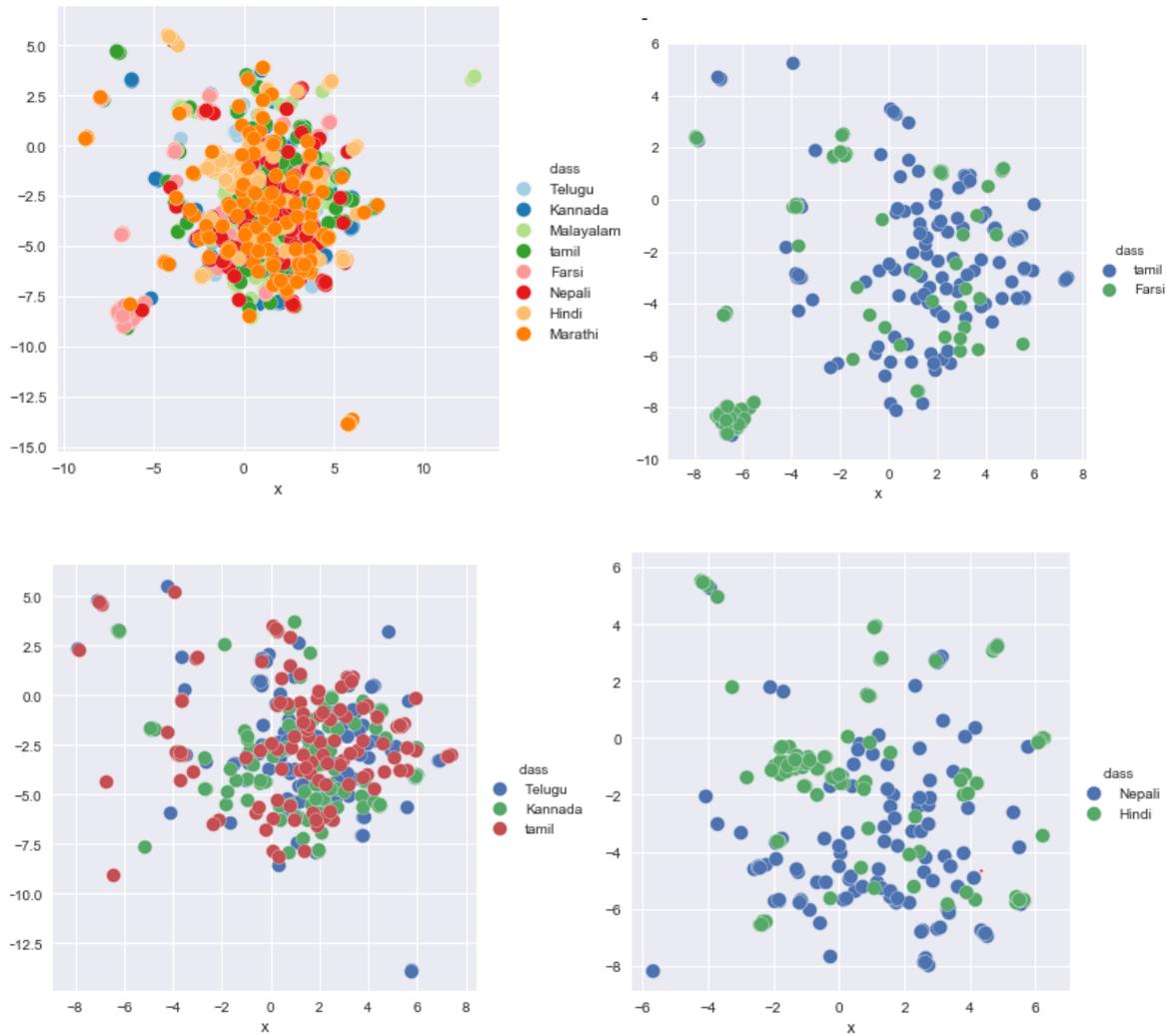

Word2Vec Embedding of the Languages

In word2vec embedding of the languages we can see a clear of a decision boundary when plotting Tamil and Farsi. Plotting Telugu, Kannada and Tamil shows significant overlap between all three languages. Same is the case for Nepalese and Hindi.

## Future Work:

These plots can be further extended to newly discovered languages and hieroglyphic scripts that were used several thousand years ago. Understanding newly discovered languages can be a challenge for linguists but is absolutely essential for archaeologists to conduct their research. We can also build deep learning model for one language and use the same model for languages with high similarity. Comparing different languages can also give us an idea of the commonly occurring stopwords with respect to every language. Hidden similarities between seemingly dissimilar languages can be uncovered and deeper analysis can even help us calculate grammatical and sentence based similarities between languages that have no characters in common

## Conclusion:

Understanding similarities between lesser known languages can help build stronger NLP models and visualizing these languages in two dimensional plots is easy to interpret. Identifying hidden language similarities can be pivotal in understanding how languages have evolved over the years and can help analyse new scripts.